\documentclass{article}

\usepackage{PRIMEarxiv}

\usepackage[utf8]{inputenc}
\usepackage[T1]{fontenc}
\usepackage{hyperref}
\usepackage{url}
\usepackage{graphicx}
\usepackage{booktabs}
\usepackage{amsfonts}
\usepackage{wrapfig}
\usepackage{threeparttable}
\usepackage{array}
\usepackage{colortbl}
\usepackage{nicefrac}
\usepackage{microtype}
\usepackage{multirow}
\usepackage{subcaption}
\usepackage{float}
\usepackage{xcolor}
\usepackage{tcolorbox}
\usepackage{amsmath}
\usepackage{enumitem}
\usepackage{pifont}
\usepackage[numbers]{natbib}

\graphicspath{{figures/}{media/}}

\definecolor{SFTcol}{HTML}{F0F8F0}
\definecolor{RLcol}{HTML}{F0F6FF}
\definecolor{SFTRLcol}{HTML}{FFF7EC}

\title{LIBERO-PRO: Towards Robust and Fair Evaluation of Vision-Language-Action Models Beyond Memorization
\thanks{\textit{\underline{Citation}}: \textbf{Zhou et al. LIBERO-PRO: Towards Robust and Fair Evaluation of Vision-Language-Action Models Beyond Memorization.}}}

\author{
{\bfseries Xueyang Zhou\textsuperscript{1}}\quad
{\bfseries Yangming Xu\textsuperscript{1}}\quad
{\bfseries Guiyao Tie\textsuperscript{1}}\quad
{\bfseries Yongchao Chen\textsuperscript{2}}\quad
{\bfseries Guowen Zhang\textsuperscript{1}}\\
{\bfseries Duanfeng Chu\textsuperscript{3}}\quad
{\bfseries Pan Zhou\textsuperscript{1}}\quad
{\bfseries Lichao Sun\textsuperscript{4}} \\
{\textsuperscript{1}Huazhong University of Science and Technology}\quad
{\textsuperscript{2}College of AI, Tsinghua University}\\
{\textsuperscript{3}Wuhan University of Technology}\quad
{\textsuperscript{4}Lehigh University}\\
\texttt{\{d202480819,U202312365,tgy,panzhou\}@hust.edu.cn}\quad
\texttt{yongchaochen12@gmail.com,lis221@lehigh.edu}
}

\begin{document}
\maketitle

\begin{abstract}
LIBERO has emerged as a widely adopted benchmark for evaluating Vision-Language-Action (VLA) models; however, its current training and evaluation settings are problematic, often leading to inflated performance estimates and preventing fair model comparison. To address these issues, we introduce LIBERO-PRO, an extended LIBERO benchmark that systematically evaluates model performance under reasonable perturbations across four dimensions: manipulated objects, initial states, task instructions, and environments. Experimental results reveal that, although existing models achieve over 90\% accuracy under the standard LIBERO evaluation, their performance collapses to 0.0\% under our generalized setting. Crucially, this discrepancy exposes the models' reliance on rote memorization of action sequences and environment layouts from the training set, rather than genuine task understanding or environmental perception. For instance, models persist in executing grasping actions when the target object is replaced with irrelevant items, and their outputs remain unchanged even when given corrupted instructions or even messy tokens. These findings expose the severe flaws in current evaluation practices, and we call on the community to abandon misleading methodologies in favor of robust assessments of model generalization and comprehension. Our code is available at: \url{https://github.com/Zxy-MLlab/LIBERO-PRO}.
\end{abstract}

\keywords{Vision-Language-Action \and Embodied AI \and Benchmark \and Robustness Evaluation}

\begin{figure*}[htb]
    \centering
    \includegraphics[width=0.999\textwidth]{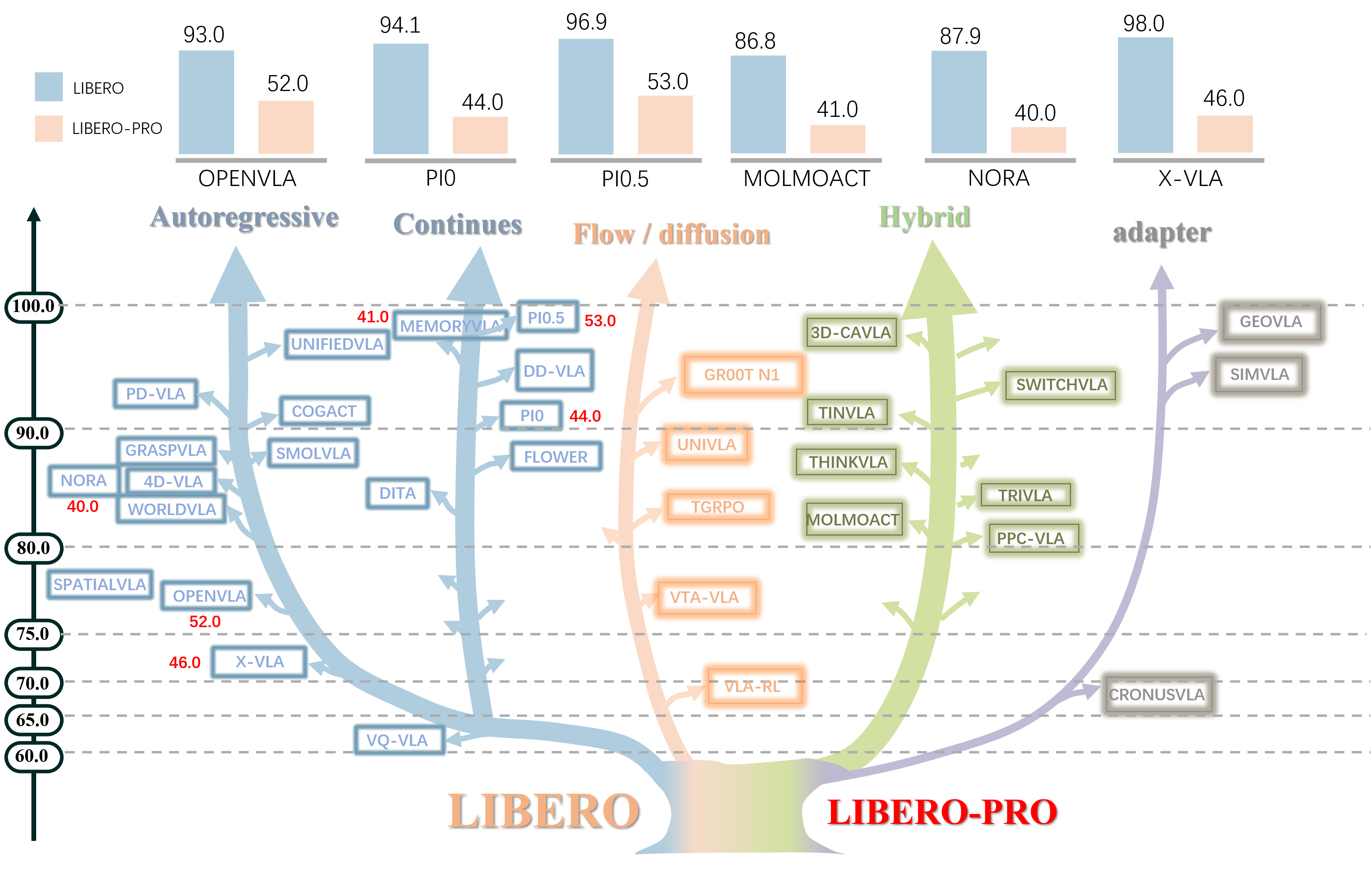}
    \caption{VLA models evaluated on LIBERO can be broadly grouped into five families according to their action decoding strategies: autoregressive, continuous, flow/diffusion, hybrid, and adapter-based methods. Although these different paradigms all achieve over 70\% performance on LIBERO, their performance drops substantially on the more challenging LIBERO-PRO benchmark. This suggests that, despite strong results on LIBERO, current VLA models still suffer from limited robustness and generalization under more demanding evaluation settings.}
    \label{fig:libero_vla}
\end{figure*}

\section{INTRODUCTION}

Vision-Language-Action (VLA) models have recently emerged as a fundamental paradigm for bridging perception, language understanding, and action execution in embodied agents~\cite{sapkota2025vision}. A central catalyst for progress in this domain is the development of standardized benchmarks~\cite{james2020rlbench,liu2023libero,mees2022calvin,li2024evaluating,nasiriany2024robocasa,walke2023bridgedata}, which enable reproducible, fair, and quantitative comparison across competing methods. Among these, LIBERO~\cite{liu2023libero} has rapidly become the most widely adopted evaluation suite for VLAs, functioning as the \emph{de facto} standard for performance reporting. As summarized in Figure~\ref{fig:libero_vla}, both large-scale pre-trained models and task-specific systems are benchmarked on LIBERO, rendering it the ``common currency'' for evaluation in this field. Consequently, the methodological rigor and reliability of LIBERO directly influence how research directions in VLA are defined and pursued.


Despite its influence, the current LIBERO training–evaluation configuration exhibits fundamental limitations. Specifically, the evaluation tasks are identical to the training tasks, differing only by marginal perturbations in initial object states—variations so subtle as to be visually imperceptible. As a result, models that overfit to the training distribution can still achieve near-perfect evaluation scores, while their accuracy collapses under even minor changes in object position or task phrasing (Figure~\ref{fig:libero_vla}). Such inflated performance fails to reflect genuine task comprehension or environmental reasoning. Our analysis further reveals that VLA models often rely on memorized task mappings rather than learned representations. Specifically, when we applied four systematic modifications—(1) \textit{replacing the target object with an unrelated one}, (2) \textit{relocating its initial position}, (3) \textit{removing it entirely}, and (4) \textit{corrupting the instruction with nonsensical tokens}—the resulting trajectories remained nearly unchanged. These findings demonstrate that, despite achieving over 90\% success on the standard LIBERO benchmark, current results primarily indicate rote recall rather than robust task understanding. \textbf{Hence, the existing evaluation protocol is insufficient to rigorously assess the reasoning, compositionality, and generalization capacity of modern VLA architectures.}

To address these shortcomings, we introduce an extended evaluation suite, \textbf{LIBERO-PRO}, designed to provide a more comprehensive and reliable measure of VLA performance. LIBERO-PRO systematically constructs evaluation scenarios along four orthogonal dimensions—manipulated objects, initial states, task instructions, and environments—while ensuring that each perturbed task remains logically coherent and physically executable. Through carefully controlled perturbations, we simulate realistic variations that better approximate the diversity of real-world conditions. Using this suite, we evaluate state-of-the-art VLA models, including OpenVLA~\cite{kim2024openvla} and Pi0/Pi0.5, and observe near-complete performance collapse under several perturbation regimes: models fail to adapt to altered object positions and exhibit severe sensitivity to even minimal instruction paraphrasing, despite encountering semantically equivalent expressions during training. These findings reveal that the widely reported accuracies exceeding 90\% largely reflect memorization rather than genuine perceptual grounding or task-level reasoning. We therefore advocate for evaluation protocols that incorporate structured perturbations, ensuring that reported progress aligns with the requirements of robust and generalizable embodied intelligence. Our key contributions are summarized as follows.

\begin{itemize}
    \item We empirically demonstrate that the current LIBERO evaluation protocol substantially overestimates model competence, as its high reported scores largely stem from memorization of training configurations rather than genuine understanding or transferable reasoning ability.
    \item We introduce LIBERO-PRO, a plug-and-play benchmark that integrates systematically designed perturbations across four core dimensions—manipulated objects, initial states, task instructions, and environments—supporting both single- and multi-dimensional randomized combinations for rigorous evaluation.
    \item We benchmark leading VLA models, including OpenVLA, Pi0, and Pi0.5, on LIBERO-PRO and reveal significant performance degradation, underscoring the limitations of the original LIBERO protocol and motivating LIBERO-PRO as a fairer, more faithful, and reproducible evaluation standard.
\end{itemize}
\section{Observing Overfitting Issues}
\label{sec:observing_overfitting_issues}
Despite most VLA models achieving over 95\% success on LIBERO, our qualitative experiments reveal several systematic failure patterns that collectively indicate severe overfitting to the training distribution. Specifically, we observe: (1) brittle visual perception, manifesting as object or location hallucinations; (2) insensitivity to instruction semantics, where actions are driven by trajectory recall rather than language grounding; and (3) a lack of compositionality, with models unable to integrate individually learned subtasks into coherent multi-step behaviors.

\subsection{Visual perceptual hallucinations}
\begin{wrapfigure}{r}{0.55\textwidth} 
    \centering
    \vspace{-25pt}
    \begin{subfigure}{0.55\textwidth}
        \includegraphics[width=\linewidth]{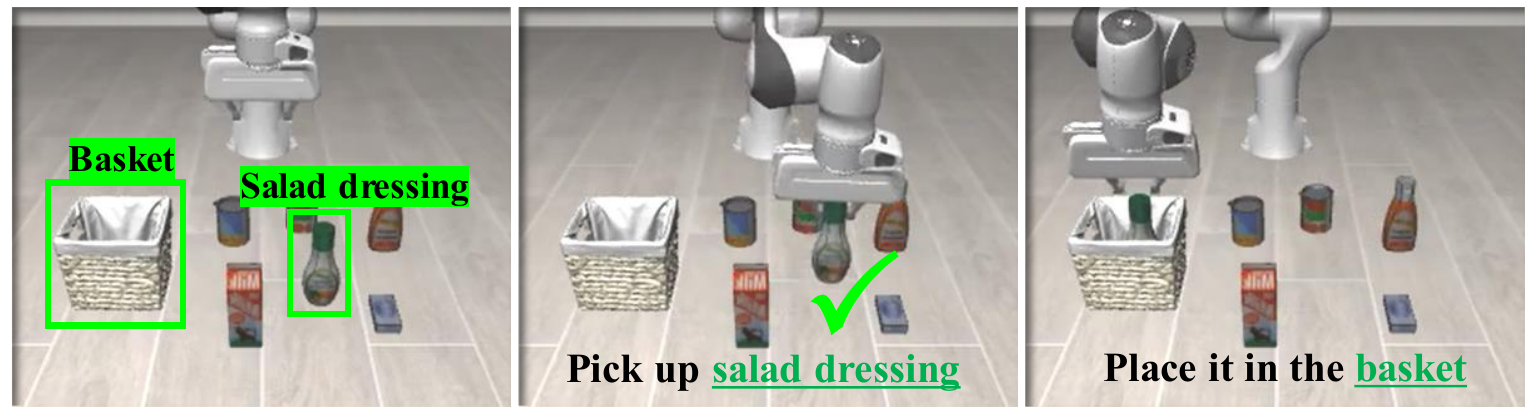}
        \caption{"salad dressing" in the training set.}
        \label{subfig:sub1}
    \end{subfigure}
    \begin{subfigure}{0.55\textwidth}
        \includegraphics[width=\linewidth]{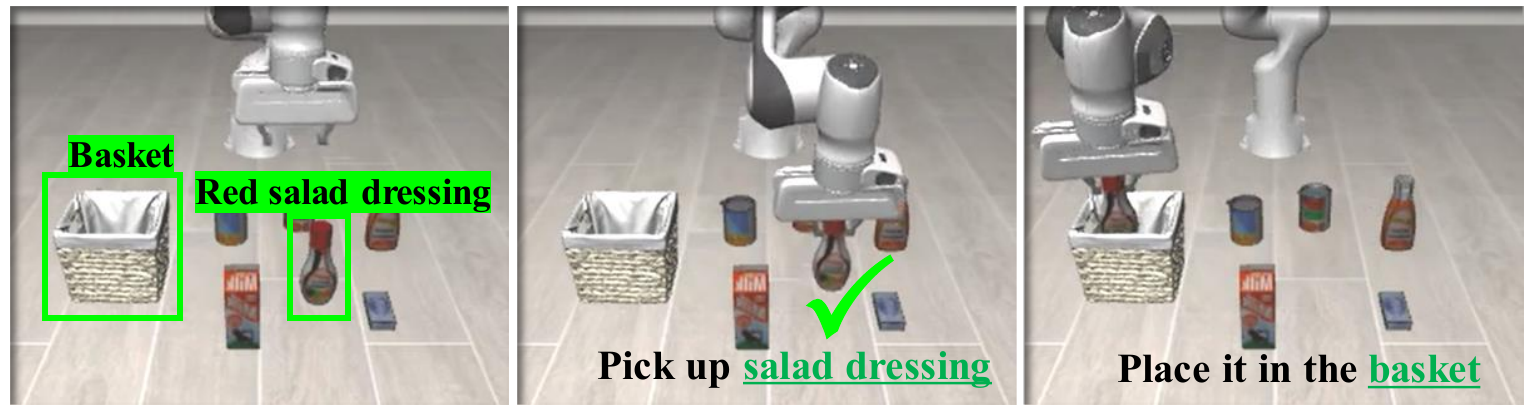}
        \caption{"salad dressing" not in the training set.}
        \label{subfig:sub2}
    \end{subfigure}
    \begin{subfigure}{0.55\textwidth}
        \includegraphics[width=\linewidth]{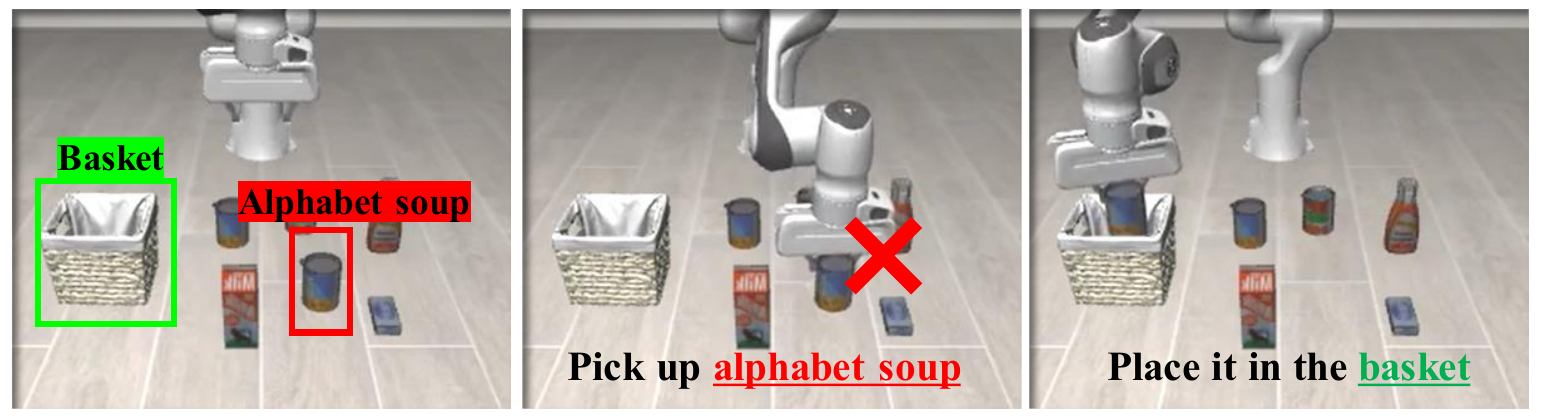}
        \caption{Replace "salad dressing" with "alphabet soup".}
        \label{subfig:sub3}
    \end{subfigure}
    \caption{Pick up salad dressing and place it in basket.}
    \vspace{-25pt}
    \label{fig:object_perturbation}
\end{wrapfigure}

\textbf{Executing the grasp on an unrelated object.} As shown in Figure \ref{fig:object_perturbation}(b), when we replace the target salad dressing bottle with an unrelated item—alphabet soup—the model still executes the instruction “Pick up the salad dressing and place it in the basket” by reaching toward and grasping the alphabet soup. This behavior indicates that the model does not reliably bind the instruction to the correct visual entity. Instead, it appears to infer the intended action from the overall scene context and subsequently replay a familiar trajectory, regardless of whether the visual evidence supports the action.

\textbf{Executing the grasp despite the object’s removal.} A similar error emerges when the target object is entirely removed from the scene. As illustrated in Figure \ref{fig:object_perturbation}(c), even though the salad dressing is absent, the model still performs the grasping motion toward the location where the object typically appears. This “grasping the air” behavior further indicates that the policy does not perform online visual verification of object existence before initiating the action.

\subsection{Instruction understand hallucinations}

\textbf{Ignoring meaningless instructions.} 
\begin{figure}[htb]  
    \centering
    \begin{subfigure}{0.48\textwidth}
        \includegraphics[width=\linewidth]{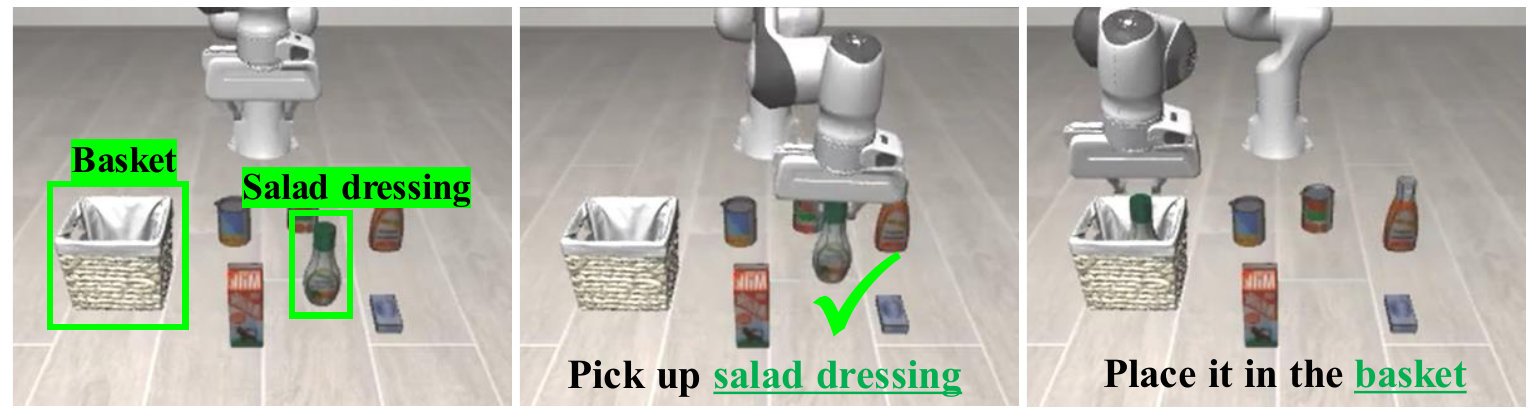}
        \caption{Original task.}
        \label{subfig:lang1}
    \end{subfigure}
    \hfill
    \begin{subfigure}{0.48\textwidth}
        \includegraphics[width=\linewidth]{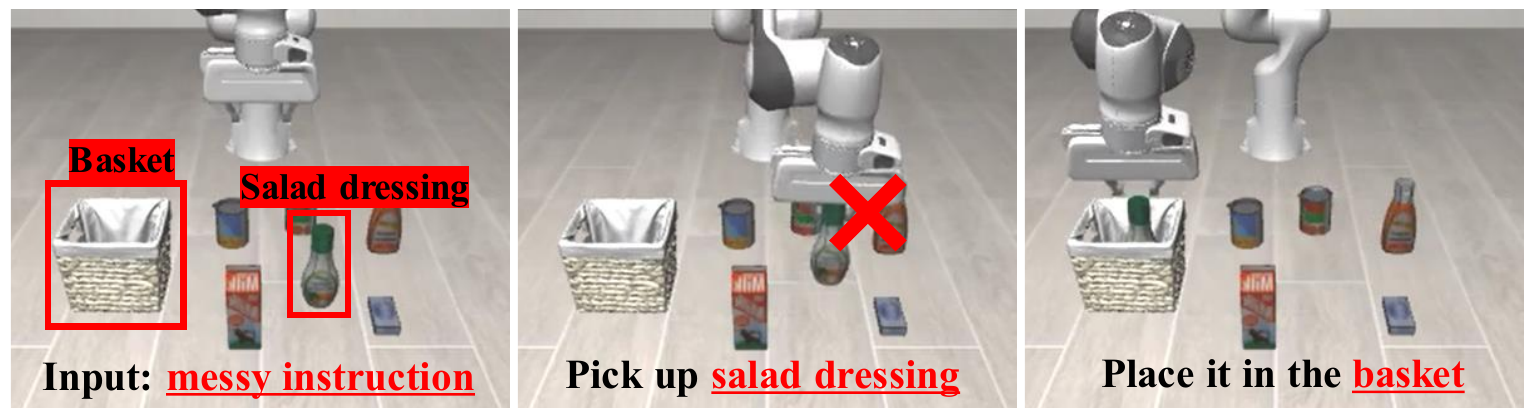}
        \caption{Meaningless task: fdsgfdsgsd or $xxx$.}
        \label{subfig:lang3}
    \end{subfigure}
    \caption{Pick up salad dressing and place it in basket.}
    \label{fig:semantic_perturbation}
\end{figure}
Figure~\ref{fig:semantic_perturbation}(b), when we replace the original instruction with a meaningless character sequence (e.g., “fdsgfdsgsd” or “xxx”), the model still executes nearly the same action trajectory as in the original task “Pick up the salad dressing and place it in the basket.” Rather than recognizing the absence of valid semantic content, the model proceeds to grasp the salad dressing and attempt the placement action without any hesitation. This behavior indicates that the policy is largely insensitive to linguistic semantics and that the instruction input contributes little to the decision-making process. Instead, the model appears to rely primarily on visual context to retrieve a memorized visuomotor pattern, further reinforcing that its apparent instruction-following capability does not stem from genuine language grounding.

\subsection{Location perception hallucinations}
\begin{wrapfigure}{r}{0.55\textwidth}
  \centering
  \vspace{-35pt}
  \centerline{\includegraphics[width=0.55\textwidth]{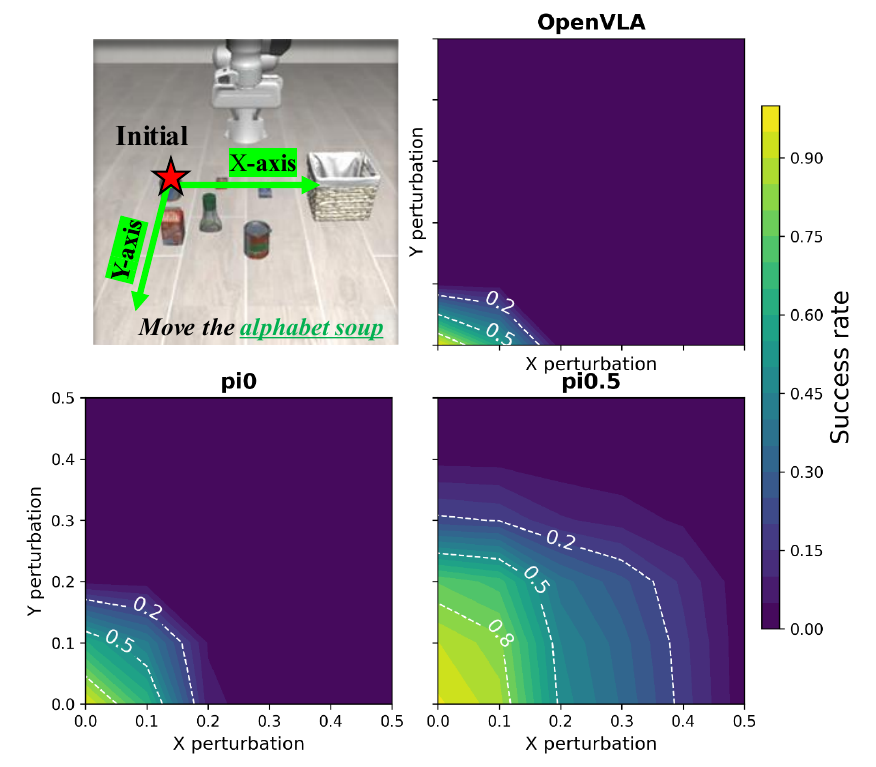}}
\caption{Success rates of OpenVLA, pi0, and pi0.5 under object position perturbations.}
\label{fig:position_perturbation}
\vspace{-35pt}
\end{wrapfigure}
Our qualitative results show that the policy is highly sensitive to the object’s initial position: when the manipulated object is displaced away from the coordinates commonly observed in the training demonstrations, the model consistently fails to execute the grasp. Even small offsets lead the end-effector to move toward the original, memorized location rather than the object’s actual position. This behavior indicates that the policy does not perform online spatial reasoning or object localization.

As shown in Figure~\ref{fig:position_perturbation}, the models are extremely sensitive to shifts in the object’s initial position: for both OpenVLA and pi0, task success collapses once the displacement exceeds 0.2 units, dropping sharply to zero thereafter. Even the stronger pi0.5 model—benefiting from a larger architecture and more diverse training data—only maintains non-trivial performance up to roughly 0.4 units before also degrading to zero.

\subsection{Lack of task understanding}
\begin{wrapfigure}{r}{0.55\textwidth}
  \centering
  \vspace{-10pt}
  \centerline{\includegraphics[width=0.55\textwidth]{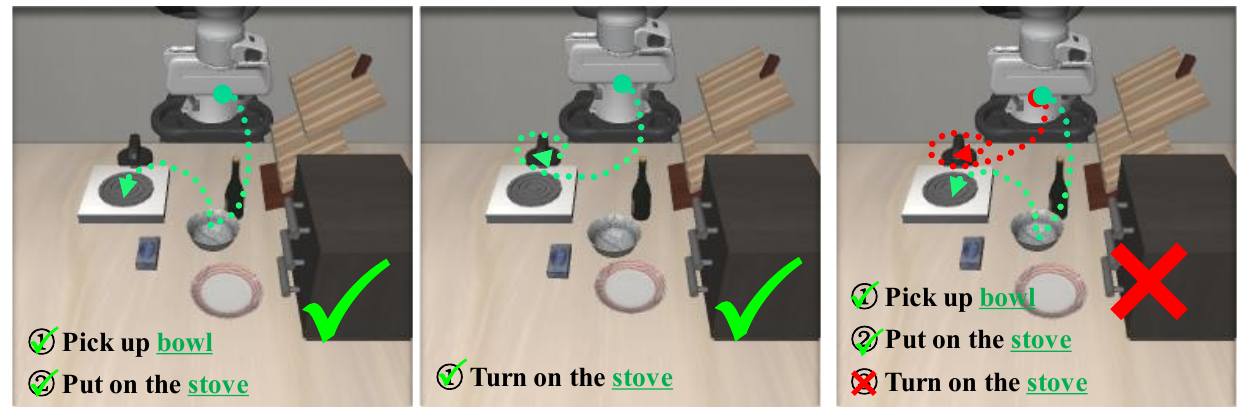}}
\caption{The model successfully executed two independent instructions, but failed to execute the combined instruction.}
\label{fig:task_perturbation}
\vspace{-10pt}
\end{wrapfigure}
To evaluate whether the model possesses genuine task-level understanding rather than merely reproducing isolated demonstrations, we design a sequence of three progressively structured tasks (Figure~\ref{fig:task_perturbation}). In the first task—pick up the bowl and place it on the stove—the model succeeds, indicating it can replicate a simple manipulation trajectory. In the second task—turn on the stove—the model again performs reliably, suggesting competence with single-step atomic actions.

However, when the two subtasks are combined into a single multi-step instruction—pick up the bowl, place it on the stove, and then turn on the stove—the model consistently fails. Rather than assembling the subtasks into a coherent execution plan, it replays partial or spurious trajectories reminiscent of the training distribution. This discrepancy reveals that the model lacks compositional reasoning: it does not understand the temporal or causal structure linking the subtasks, but instead depends on memorized action fragments. Even such modest task compositions exceed its generalization capacity, highlighting the fragility of current VLA models.



These observations suggest that the strong in-distribution performance of current Visual-Language-Action (VLA) models may rely heavily on superficial regularities in the training data. Although a VLA policy is intended to learn a mapping $\pi:(\mathcal{O},\mathcal{L})\rightarrow\mathcal{A}$ from visual observations and language instructions to actions, our results indicate that the learned policy often fails to robustly ground actions in the actual observation-instruction pair. Instead, it appears to exploit shortcuts such as canonical scene layouts, recurring object appearances, typical object positions, and repeated motion trajectories. This provides a plausible explanation for the observed failures: the model grasps absent or incorrect objects, ignores meaningless instructions, reaches toward memorized locations under positional perturbations, and fails to compose individually learned subtasks into coherent multi-step behaviors. Overall, these results suggest that current VLA models may still depend more on shortcut feature-action correlations than on robust and compositional visuomotor understanding.
\section{Problem Definition}

This section formalizes the Vision-Language-Action (VLA) task framework and examines limitations in the LIBERO benchmark’s evaluation protocol that may obscure true task competence.

\subsection{Formal VLA Task Definition}

We formalize a vision-language-action (VLA) task instance as
\begin{equation}
\mathcal{T} = (l, e, G),
\end{equation}
where \(l \in \mathcal{L}\) is a natural-language instruction, \(e \in \mathcal{E}\) denotes an environment instance, \(\mathcal{E}\) is the space of environment instances, and \(G : \mathcal{S} \times \mathcal{L} \to \{0,1\}\) is a binary success predicate indicating whether a state satisfies the task specified by the instruction. An environment instance is defined as
\begin{equation}
e = (\mathcal{W}, \mathcal{O}, R_0),
\end{equation}
where \(\mathcal{W}\) denotes the visual environment context (e.g., background, textures, and lighting conditions), \(\mathcal{O}\) denotes the set of interactive objects in the scene, and \(R_0\) denotes the initial spatial configuration of the objects. Let \(s_0\) denote an initial state consistent with \(R_0\). A policy \(\pi_\theta(a_t \mid o_t, l)\) induces trajectories according to
\begin{equation}
a_t \sim \pi_\theta(\cdot \mid o_t, l), \qquad
s_{t+1} \sim P(\cdot \mid s_t, a_t),
\end{equation}
where \(o_t\) denotes the agent observation at time \(t\), and \(P : \mathcal{S} \times \mathcal{A} \to \Delta(\mathcal{S})\) denotes the environment dynamics. We evaluate a policy by its success probability within a finite horizon \(H\):
\begin{equation}
J(\pi_\theta; \mathcal{T})
=
\mathbb{P}_{\pi_\theta,\, e}\!\left[\exists t \in \{0,\dots,H\},\; G(s_t, l)=1\right],
\end{equation}
where the probability is taken over trajectories induced by \(\pi_\theta\) under the environment \(e\). This objective captures the joint requirements of language grounding, perception, and action.

\subsection{Current LIBERO Evaluation Protocol}

While LIBERO has been an influential benchmark for VLA research, its current evaluation protocol exhibits limited distributional shift between training and test tasks. Under our formulation, each task is specified by an instruction \(l\), an environment instance \(e = (\mathcal{W}, \mathcal{O}, R_0)\), and a success predicate \(G\).

In LIBERO, a test task \(\mathcal{T}' \in \mathcal{T}_{\mathrm{test}}\) is typically constructed from a corresponding training task \(\mathcal{T} \in \mathcal{T}_{\mathrm{train}}\) such that
\begin{equation}
l' = l, \qquad
\mathcal{W}' = \mathcal{W}, \qquad
\mathcal{O}' = \mathcal{O}, \qquad
G' = G,
\end{equation}
with variation arising only from small perturbations to the initial spatial configuration \(R_0\), i.e., \(R_0'\) differs from \(R_0\) only by minor perturbations.

Consequently, test tasks remain close to training tasks, and evaluation primarily reflects robustness to small variations in object configurations rather than generalization to novel task compositions or environments. As a result, strong performance may largely arise from interpolation within a narrow neighborhood of the training distribution, potentially overestimating a model's ability to generalize under broader distribution shifts.
\section{Methodology}

\subsection{Proposed Perturbation-Based Evaluation Framework}

\begin{figure*}[t]
    \centering
    \includegraphics[width=0.999\textwidth]{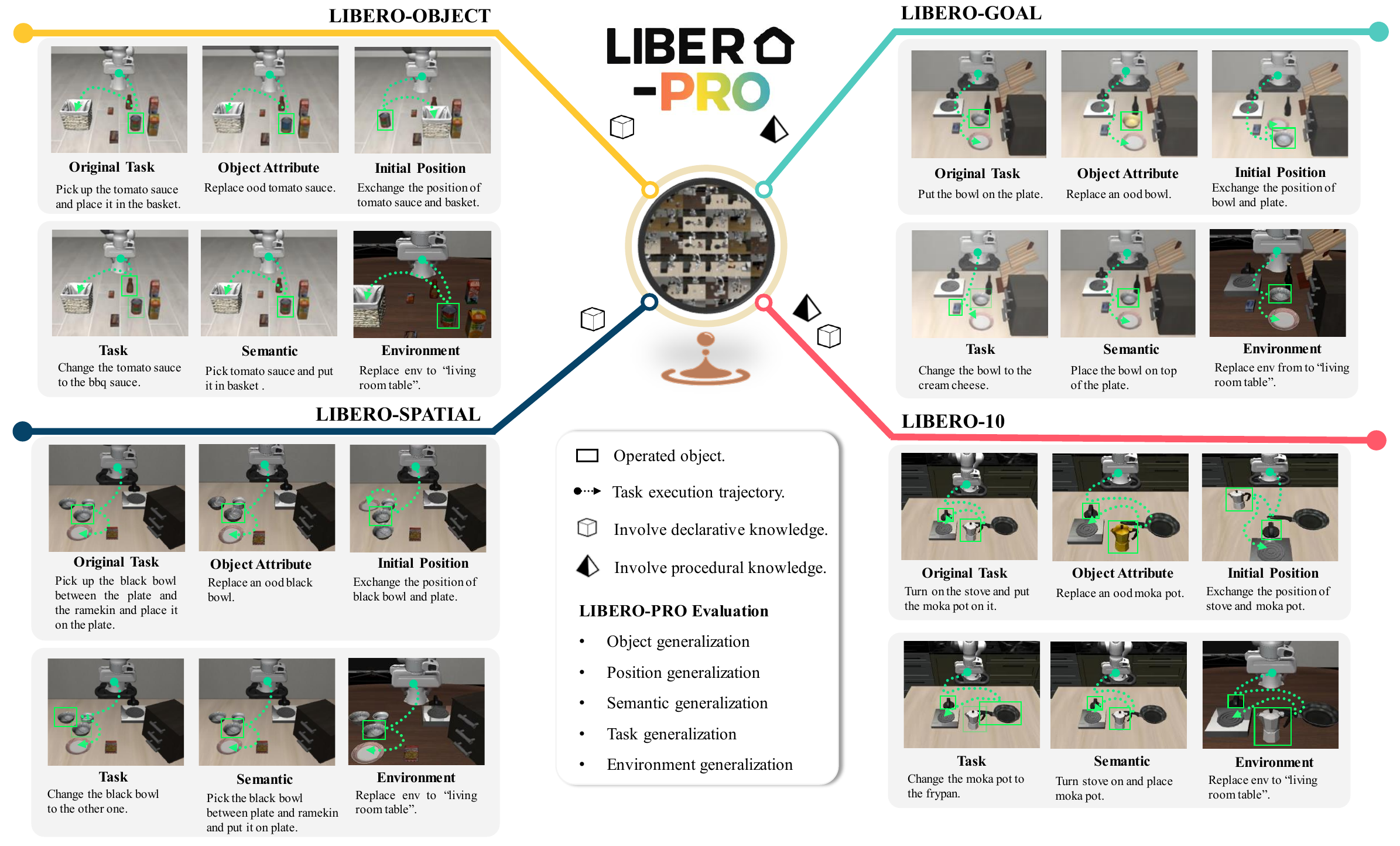}
    \caption{Overview of LIBERO-PRO. LIBERO-PRO extends LIBERO with five perturbation types: object, configuration, instruction, task, and environment. These perturbations vary object attributes, object placements, language descriptions, task goals/object sets, and visual context, respectively, enabling systematic evaluation under diverse distribution shifts. To ensure feasibility, task perturbations are excluded from cross-type combinations, while the other perturbation types can be freely composed.}
    \label{fig:libero_vla}
\end{figure*}

To enable more principled evaluation of VLA models, we design a perturbation-based framework that introduces controlled variations across four orthogonal dimensions: object attributes, initial spatial configurations, task instructions, and environments. Given a task \(\mathcal{T}=(l,e,G)\) with environment instance \(e=(\mathcal{W},\mathcal{O},R_0)\), we generate perturbed variants \(\mathcal{T}^{(k)}=\phi_k(\mathcal{T})\), where \(k\in\{\mathrm{O},\mathrm{R},\mathrm{L},\mathrm{E}\}\).

\textbf{Object attribute perturbation} (\(k=\mathrm{O}\)) modifies non-essential visual attributes of objects, such as color, texture, or size, while preserving semantic equivalence. The resulting task is
\[
\mathcal{T}^{(\mathrm{O})} = (l, e^{(\mathrm{O})}, G), 
\qquad
e^{(\mathrm{O})} = (\mathcal{W}, \mathcal{O}^{(\mathrm{O})}, R_0).
\]

\textbf{Initial configuration perturbation} (\(k=\mathrm{R}\)) alters the absolute and relative positions of objects while maintaining physical plausibility. The perturbed task is
\[
\mathcal{T}^{(\mathrm{R})} = (l, e^{(\mathrm{R})}, G), 
\qquad
e^{(\mathrm{R})} = (\mathcal{W}, \mathcal{O}, R_0^{(\mathrm{R})}).
\]

\textbf{Instruction perturbation} (\(k=\mathrm{L}\)) includes both semantic- and task-level variations. Semantic perturbations rephrase the instruction while preserving task intent:
\[
\mathcal{T}^{(\mathrm{L}_{\mathrm{sem}})} = (l^{(\mathrm{L})}, e, G).
\]

\textbf{Task perturbation} (\(k=\mathrm{T}\)) modifies the target object or action while remaining within the training distribution. The perturbed task is
\[
\mathcal{T}^{(\mathrm{T})} = (l, e, G^{(\mathrm{T})}).
\]

\textbf{Environment perturbation} (\(k=\mathrm{E}\)) modifies the visual environment context, such as background, lighting, or texture, without altering task feasibility. The resulting task is
\[
\mathcal{T}^{(\mathrm{E})} = (l, e^{(\mathrm{E})}, G), 
\qquad
e^{(\mathrm{E})} = (\mathcal{W}^{(\mathrm{E})}, \mathcal{O}, R_0).
\]

\subsection{Perturbation Constraints}

To ensure that each perturbed task remains semantically coherent and executable, we impose two classes of constraints. First, perturbations are bounded within a dimension-specific neighborhood:
\[
x^{(k)} \in \mathcal{N}_k(x), 
\qquad
\mathrm{dist}_k(x, x^{(k)}) \leq \delta_k,
\]
where \(x\) denotes a task component, \(x^{(k)}\) its perturbed counterpart, and \(\delta_k\) controls the perturbation magnitude along dimension \(k\). Second, we enforce non-trivial task-level separation:
\[
d_{\mathrm{TV}}(\mathcal{T}, \mathcal{T}^{(k)}) > \epsilon,
\]
where \(d_{\mathrm{TV}}\) denotes the total variation distance, ensuring that perturbations induce meaningful behavioral changes. By varying \(\delta_k\), we generate multiple perturbed variants for each category and characterize policy sensitivity to perturbation intensity.

\subsection{LIBERO-PRO Evaluation Benchmark}

Building upon the above framework, we introduce \textbf{LIBERO-PRO}, a plug-and-play benchmark for systematic evaluation across multiple dimensions of variation, including object attributes, spatial configurations, instructions, tasks, and environments, as well as their combinations.

For \textit{object variation}, we modify object color, texture, and scale. For \textit{configuration variation}, we alter object placements while preserving physical validity. For \textit{instruction variation}, we generate paraphrased task descriptions. For \textit{task variation}, we redesign tasks to introduce new object sets and goals while maintaining executability. For \textit{environment variation}, we vary visual context such as backgrounds and lighting.

To avoid infeasible combinations, task-level perturbations are excluded from cross-type compositions, while the remaining perturbation types may be freely combined to construct diverse evaluation settings.
\section{Experiments}

\subsection{Settings}

\begin{figure*}[htb]
    \centering
    \includegraphics[width=0.999\textwidth]{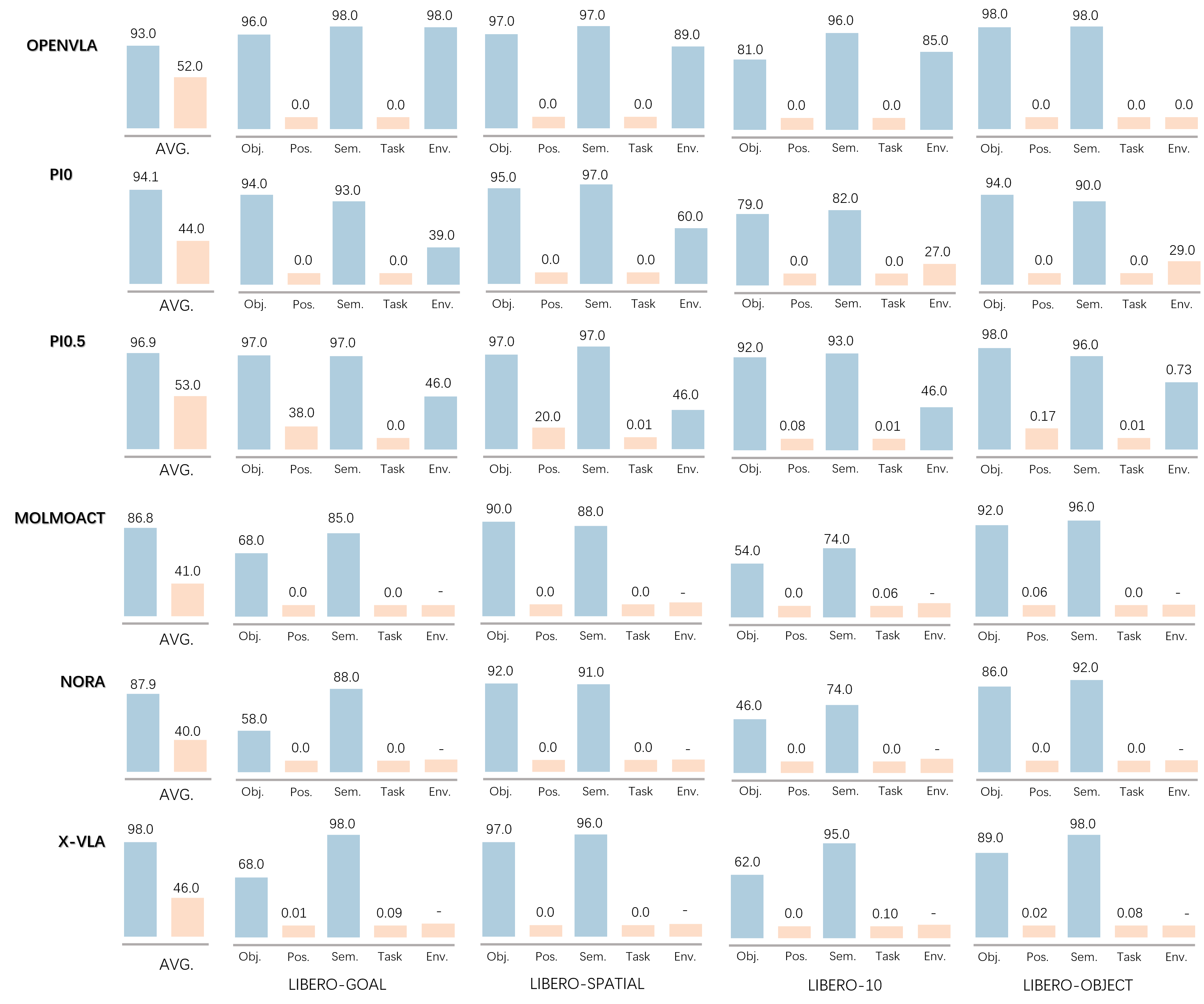}
    \caption{Main results on LIBERO-PRO. Despite high average success on standard LIBERO tasks, current VLA models fail sharply under controlled perturbations, particularly position and task shifts. Robustness to object, semantic instruction, and environment changes is limited and varies across models and task suites.}
    \label{fig:preformance}
\end{figure*}
We conduct experiments on LIBERO-PRO, which applies controlled perturbations from four perspectives: object properties, initial positions, task instructions (covering both semantic and structural variations), and environments. We evaluate three representative models—OpenVLA\cite{kim2024openvla}, pi0\cite{black2024pi0}, and pi0.5\cite{intelligence2025pi05visionlanguageactionmodelopenworld}. For fairness, OpenVLA is assessed using four official checkpoints trained on different LIBERO suites, whereas pi0 and pi0.5 are evaluated with a single official checkpoint. Consistent with the original LIBERO protocol, we set the number of evaluation episodes to 50 per task.

\subsection{Main Results}
We conducted a comprehensive evaluation of state-of-the-art vision-language-action (VLA) models \cite{black2024pi0,kim2024openvla} using the LIBERO-PRO evaluation suite, which introduces perturbations across four dimensions: objects, goals, spatial configurations, and task sets (see Tables \ref{fig:preformance}). Our findings are as follows: (1) \textbf{Failure under simple perturbations.} Despite achieving success rates above 90\% on the standard LIBERO benchmark, models nearly collapse under changes to object positions or minor task modifications, even when constructed from training components. (2) \textbf{Illusory robustness.} Apparent stability against object and instruction perturbations is largely due to overfitting rather than genuine task understanding. (3) \textbf{Model-specific sensitivity.} Environmental perturbations affect models differently, depending on the degree of environmental divergence and reliance on memorized contexts. (4) \textbf{Subtle capability differences.} While most models fail under perturbations, Pi0.5 achieves a 0.38 success rate in the libero-goal task under position changes, compared to 0 for OpenVLA and Pi0, revealing differences masked by standard benchmark scores. Overall, these results show that current VLA models, despite excelling on the standard LIBERO, lack true robustness and generalization—highlighting the need for rigorous evaluation frameworks.

\section{Related Work}

\noindent\textbf{Vision-Language-Action Models.}
Vision-Language-Action (VLA) models have rapidly emerged as a central paradigm for embodied AI, aiming to unify perception, natural language understanding, and action execution within a single framework. Recent years have witnessed an explosion of approaches, ranging from large-scale pre-trained models that leverage multimodal corpora to task-specific architectures tailored for robotic manipulation and navigation\cite{ma2025surveyvisionlanguageactionmodelsembodied,li2025surveyvisionlanguageactionmodelsembodied,din2025visionlanguageactionmodels}. While these efforts have significantly advanced the field, they also underscore the urgent need for standardized evaluation benchmarks\cite{liu2023libero, mees2022calvin, james2020rlbench}. Without such benchmarks, it becomes difficult to conduct fair and reproducible comparisons across methods, hindering the cumulative progress of VLA research.

\noindent\textbf{Benchmarks for VLA Evaluation.}
A number of benchmarks have been developed to enable systematic evaluation of VLA models, including RLBench \cite{james2020rlbench}, CALVIN \cite{mees2022calvin}, RoboCasa \cite{nasiriany2024robocasa}, and BridgeData \cite{walke2023bridgedata}. These benchmarks provide valuable testbeds for task execution under vision and language guidance, yet their adoption has remained relatively limited within the broader community. In contrast, LIBERO \cite{liu2023libero} has quickly emerged as the dominant benchmark and is now the most widely adopted protocol for evaluating VLA models. Its standardized task suite and unified reporting metrics have made it the de facto standard for performance comparison, to the extent that virtually all recent VLA studies report results on LIBERO. As such, LIBERO not only serves as the “common currency” of evaluation in this field, but also exerts a strong influence on research directions and claims of progress.

\noindent\textbf{Limitations of Current Evaluation Practices.}
Several recent studies have noted that the high scores reported on LIBERO often fail to translate into reliable performance, even on tasks with only minor variations. This has raised serious concerns about the validity of its evaluation protocol\cite{zhou2025exploringlimitsvisionlanguageactionmanipulations,fang2025intentionexecutionprobinggeneralization, li2025taskreconstructionextrapolationpi0}. In response, some researchers have explored stronger algorithms or proposed stress tests \cite{gu2025safemultitaskfailuredetection, guruprasad2024benchmarkingvisionlanguage, valle2025evaluatinguncertaintyqualityvisual, kube2025beyond,wang2025exploringadversarialvulnerabilitiesvisionlanguageaction}to encourage more robust VLA systems. However, the field still lacks a systematic and standardized framework for robust evaluation under LIBERO. The majority of works continue to rely on the original, flawed setup, reporting numbers that may not reflect genuine capability. To address this gap, we propose LIBERO-PRO, a plug-and-play extension of LIBERO designed to foster fair comparison across models and to advance research on building VLA systems that can withstand reasonable and practically necessary perturbations.
\section{Conclusion}
In conclusion, our study reveals that the current LIBERO evaluation protocol is fundamentally flawed: by reusing identical training and evaluation tasks with only imperceptible perturbations, it fails to measure genuine capability. Reported accuracies above 90\% largely reflect rote memorization of fixed mappings from the training set, rather than true comprehension of instructions or acquisition of robust action strategies. We call on the community to move beyond misleading evaluation practices and adopt our LIBERO-PRO benchmark, which provides a principled and effective framework for assessing the real generalization and understanding capabilities of VLA models.

\bibliographystyle{plainnat}
\bibliography{references}

@article{kim2024openvla,
  title={Openvla: An open-source vision-language-action model},
  author={Kim, M J and Pertsch, K and Karamcheti, S and others},
  journal={arXiv preprint arXiv:2406.09246},
  year={2024}
}

@article{black2024pi0,
  title={$\pi_0$: A Vision-Language-Action Flow Model for General Robot Control},
  author={Black, K and Brown, N and Driess, D and others},
  journal={arXiv preprint arXiv:2410.24164},
  year={2024}
}

@article{james2020rlbench,
  title={Rlbench: The robot learning benchmark \& learning environment},
  author={James, Stephen and Ma, Zicong and Arrojo, David Rovick and Davison, Andrew J},
  journal={IEEE Robotics and Automation Letters},
  volume={5},
  number={2},
  pages={3019--3026},
  year={2020},
  publisher={IEEE}
}

@article{li2024evaluating,
  title={Evaluating real-world robot manipulation policies in simulation},
  author={Li, Xuanlin and Hsu, Kyle and Gu, Jiayuan and Pertsch, Karl and Mees, Oier and Walke, Homer Rich and Fu, Chuyuan and Lunawat, Ishikaa and Sieh, Isabel and Kirmani, Sean and others},
  journal={arXiv preprint arXiv:2405.05941},
  year={2024}
}

@article{liu2023libero,
  title={Libero: Benchmarking knowledge transfer for lifelong robot learning},
  author={Liu, Bo and Zhu, Yifeng and Gao, Chongkai and Feng, Yihao and Liu, Qiang and Zhu, Yuke and Stone, Peter},
  journal={Advances in Neural Information Processing Systems},
  volume={36},
  pages={44776--44791},
  year={2023}
}

@article{mees2022calvin,
  title={Calvin: A benchmark for language-conditioned policy learning for long-horizon robot manipulation tasks},
  author={Mees, Oier and Hermann, Lukas and Rosete-Beas, Erick and Burgard, Wolfram},
  journal={IEEE Robotics and Automation Letters},
  volume={7},
  number={3},
  pages={7327--7334},
  year={2022},
  publisher={IEEE}
}

@article{nasiriany2024robocasa,
  title={Robocasa: Large-scale simulation of everyday tasks for generalist robots},
  author={Nasiriany, Soroush and Maddukuri, Abhiram and Zhang, Lance and Parikh, Adeet and Lo, Aaron and Joshi, Abhishek and Mandlekar, Ajay and Zhu, Yuke},
  journal={arXiv preprint arXiv:2406.02523},
  year={2024}
}

@inproceedings{walke2023bridgedata,
  title={Bridgedata v2: A dataset for robot learning at scale},
  author={Walke, Homer Rich and Black, Kevin and Zhao, Tony Z and Vuong, Quan and Zheng, Chongyi and Hansen-Estruch, Philippe and He, Andre Wang and Myers, Vivek and Kim, Moo Jin and Du, Max and others},
  booktitle={Conference on Robot Learning},
  pages={1723--1736},
  year={2023},
  organization={PMLR}
}

@article{sapkota2025vision,
  title={Vision-language-action models: Concepts, progress, applications and challenges},
  author={Sapkota, Ranjan and Cao, Yang and Roumeliotis, Konstantinos I and Karkee, Manoj},
  journal={arXiv preprint arXiv:2505.04769},
  year={2025}
}

@misc{ma2025surveyvisionlanguageactionmodelsembodied,
      title={A Survey on Vision-Language-Action Models for Embodied AI}, 
      author={Yueen Ma and Zixing Song and Yuzheng Zhuang and Jianye Hao and Irwin King},
      year={2025},
      eprint={2405.14093},
      archivePrefix={arXiv},
      primaryClass={cs.RO},
      url={https://arxiv.org/abs/2405.14093}, 
}

@misc{li2025surveyvisionlanguageactionmodelsembodied,
      title={Survey of Vision-Language-Action Models for Embodied Manipulation}, 
      author={Haoran Li and Yuhui Chen and Wenbo Cui and Weiheng Liu and Kai Liu and Mingcai Zhou and Zhengtao Zhang and Dongbin Zhao},
      year={2025},
      eprint={2508.15201},
      archivePrefix={arXiv},
      primaryClass={cs.RO},
      url={https://arxiv.org/abs/2508.15201}, 
}

@misc{din2025visionlanguageactionmodels,
      title={Vision Language Action Models in Robotic Manipulation: A Systematic Review}, 
      author={Muhayy Ud Din and Waseem Akram and Lyes Saad Saoud and Jan Rosell and Irfan Hussain},
      year={2025},
      eprint={2507.10672},
      archivePrefix={arXiv},
      primaryClass={cs.RO},
      url={https://arxiv.org/abs/2507.10672}, 
}

@misc{intelligence2025pi05visionlanguageactionmodelopenworld,
      title={$\pi_{0.5}$: a Vision-Language-Action Model with Open-World Generalization}, 
      author={Physical Intelligence and Kevin Black and Noah Brown and James Darpinian and Karan Dhabalia and Danny Driess and Adnan Esmail and Michael Equi and Chelsea Finn and Niccolo Fusai and Manuel Y. Galliker and Dibya Ghosh and Lachy Groom and Karol Hausman and Brian Ichter and Szymon Jakubczak and Tim Jones and Liyiming Ke and Devin LeBlanc and Sergey Levine and Adrian Li-Bell and Mohith Mothukuri and Suraj Nair and Karl Pertsch and Allen Z. Ren and Lucy Xiaoyang Shi and Laura Smith and Jost Tobias Springenberg and Kyle Stachowicz and James Tanner and Quan Vuong and Homer Walke and Anna Walling and Haohuan Wang and Lili Yu and Ury Zhilinsky},
      year={2025},
      eprint={2504.16054},
      archivePrefix={arXiv},
      primaryClass={cs.LG},
      url={https://arxiv.org/abs/2504.16054}, 
}

@misc{li2025taskreconstructionextrapolationpi0,
      title={Task Reconstruction and Extrapolation for $\pi_0$ using Text Latent}, 
      author={Quanyi Li},
      year={2025},
      eprint={2505.03500},
      archivePrefix={arXiv},
      primaryClass={cs.RO},
      url={https://arxiv.org/abs/2505.03500}, 
}

@misc{fang2025intentionexecutionprobinggeneralization,
      title={From Intention to Execution: Probing the Generalization Boundaries of Vision-Language-Action Models}, 
      author={Irving Fang and Juexiao Zhang and Shengbang Tong and Chen Feng},
      year={2025},
      eprint={2506.09930},
      archivePrefix={arXiv},
      primaryClass={cs.RO},
      url={https://arxiv.org/abs/2506.09930}, 
}

@misc{zhou2025exploringlimitsvisionlanguageactionmanipulations,
      title={Exploring the Limits of Vision-Language-Action Manipulations in Cross-task Generalization}, 
      author={Jiaming Zhou and Ke Ye and Jiayi Liu and Teli Ma and Zifan Wang and Ronghe Qiu and Kun-Yu Lin and Zhilin Zhao and Junwei Liang},
      year={2025},
      eprint={2505.15660},
      archivePrefix={arXiv},
      primaryClass={cs.RO},
      url={https://arxiv.org/abs/2505.15660}, 
}

@misc{gu2025safemultitaskfailuredetection,
      title={SAFE: Multitask Failure Detection for Vision-Language-Action Models}, 
      author={Qiao Gu and Yuanliang Ju and Shengxiang Sun and Igor Gilitschenski and Haruki Nishimura and Masha Itkina and Florian Shkurti},
      year={2025},
      eprint={2506.09937},
      archivePrefix={arXiv},
      primaryClass={cs.RO},
      url={https://arxiv.org/abs/2506.09937}, 
}

@misc{guruprasad2024benchmarkingvisionlanguage,
      title={Benchmarking Vision, Language, \& Action Models on Robotic Learning Tasks}, 
      author={Pranav Guruprasad and Harshvardhan Sikka and Jaewoo Song and Yangyue Wang and Paul Pu Liang},
      year={2024},
      eprint={2411.05821},
      archivePrefix={arXiv},
      primaryClass={cs.RO},
      url={https://arxiv.org/abs/2411.05821}, 
}

@misc{valle2025evaluatinguncertaintyqualityvisual,
      title={Evaluating Uncertainty and Quality of Visual Language Action-enabled Robots}, 
      author={Pablo Valle and Chengjie Lu and Shaukat Ali and Aitor Arrieta},
      year={2025},
      eprint={2507.17049},
      archivePrefix={arXiv},
      primaryClass={cs.SE},
      url={https://arxiv.org/abs/2507.17049}, 
}

@article{kube2025beyond,
  title={Beyond performance: Explaining generalisation failures of Robotic Foundation Models in industrial simulation},
  author={Kube, David and Hadwiger, Simon and Meisen, Tobias},
  journal={Biomimetic Intelligence and Robotics},
  pages={100249},
  year={2025},
  publisher={Elsevier}
}

@misc{wang2025exploringadversarialvulnerabilitiesvisionlanguageaction,
      title={Exploring the Adversarial Vulnerabilities of Vision-Language-Action Models in Robotics}, 
      author={Taowen Wang and Cheng Han and James Chenhao Liang and Wenhao Yang and Dongfang Liu and Luna Xinyu Zhang and Qifan Wang and Jiebo Luo and Ruixiang Tang},
      year={2025},
      eprint={2411.13587},
      archivePrefix={arXiv},
      primaryClass={cs.RO},
      url={https://arxiv.org/abs/2411.13587}, 
}

\end{document}